# Applying Large Language Models and Chain-of-Thought for Automatic Scoring

Gyeong-Geon Lee[1], Ehsan Latif[1], Xuansheng Wu[2], Ninghao Liu[2], Xiaoming Zhai[1*]

[1] *AI4STEM Education Center, University of Georgia, Athens, GA, USA;*

[2] *School of Computing, University of Georgia, Athens, GA, USA*

*corresponding author: xiaoming.zhai@uga.edu

**Abstract**

This study investigates the application of large language models (LLMs), specifically GPT-3.5 and GPT-4, with Chain-of-Though (CoT) in the automatic scoring of student-written responses to science assessments. We focused on overcoming the challenges of accessibility, technical complexity, and lack of explainability that have previously limited the use of artificial intelligence-based automatic scoring tools among researchers and educators. With a testing dataset comprising six assessment tasks (three binomial and three trinomial) with 1,650 student responses, we employed six prompt engineering strategies to automatically score student responses. The six strategies combined zero-shot or few-shot learning with CoT, either alone or alongside item stem and scoring rubrics, developed based on a novel approach, WRVRT (prompt writing, reviewing, validating, revising, and testing). Results indicated that few-shot (acc = .67) outperformed zero-shot learning (acc = .60), with 12.6% increase. CoT, when used without item stem and scoring rubrics, did not significantly affect scoring accuracy (acc = .60). However, CoT prompting paired with contextual item stems and rubrics proved to be a significant contributor to scoring accuracy (13.44% increase for zero-shot; 3.7% increase for few-shot). We found a more balanced accuracy across different proficiency categories when CoT was used with a scoring rubric, highlighting the importance of domain-specific reasoning in enhancing the effectiveness of LLMs in scoring tasks. We also found that GPT-4 demonstrated superior performance over GPT -3.5 in various scoring tasks when combined with the single-call greedy sampling or ensemble voting nucleus sampling strategy, showing 8.64% difference. Particularly, the single-call greedy sampling strategy with GPT-4 outperformed other approaches. This study also demonstrates the potential of LLMs in facilitating explainable and interpretable automatic scoring, emphasizing that CoT enhances accuracy and transparency, particularly when used with item stem and scoring rubrics.

# 1 Introduction

The field of education is undergoing a transformation with the increasing integration of artificial intelligence (AI) to enhance teaching and learning. Within this transformative landscape, automatic scoring systems have emerged as indispensable tools. They play a pivotal role in meeting the pressing need for efficient, precise, and timely assessment of students' proficiency in applying knowledge to solve problems (Zhai, Haudek et al., 2020). While automatic scoring can be applied for many subject matters, science education is in particular need for such systems to address the expansive scope of the integrated science disciplines, engage students in solving real-world problems, and advance the intricate nature of assessment practices for learning. Automatic scoring enables immediate feedback, which is crucial for fostering an adaptive learning environment where students can promptly recognize and rectify misunderstandings, thus enhancing their use of disciplinary core ideas and crosscutting concepts to solve problems (Zhai, 2021).

Existing methods of automatic scoring have largely hinged on the advancements in machine learning and natural language processing (NLP). Techniques ranging from individual algorithms (Nehm, Ha, & Mayfield, 2012), ensemble algorithms that utilizes multiple scoring models rather than a single model (Wilson et al., 2023), to sophisticated large language models (LLMs) (Latif & Zhai, 2023; Z. Liu, et al., 2023) have been employed to evaluate short-answer questions to extensive essays. These systems have made strides in understanding the syntactical structure of student responses but frequently grapple with the nuances of scientific reasoning and the interpretation of students' thinking processes. Despite the progress, research suggests that developing such scoring models is time- and effort-consuming (Zhai, in press). Therefore, recent studies leverage prompt engineering and have reported the possibility of leveraging this new method to free researchers from labelling a large number of training cases (Wu et al., 2023). However, the reported scoring accuracy needs significant efforts to improve, which is often attributed to the limited capacity of LLMs to grasp the depth of content-specific knowledge and the rationale behind students' answers. Also, while many LLMs, such as the Generative Pre-trained Transformer (GPT) family, have been released until now, the question of which models and which settings of hyperparameters could best serve the automatic scoring has not been answered. If Gemini Pro by Google shows less performance than GPT-4(V) for educational tasks (Lee, Latif et al., 2023), it is recommended to further explore how to fully exploit GPT variants' function for automatic scoring.

To address these research gaps, this study posits that the integration of LLMs with chain-of-thought (CoT) prompting methods could significantly enhance the accuracy of automatic scoring systems in science education. CoT is characterized as a sequence of intermediary reasoning steps expressed in natural language, culminating in the final output (Wei et al., 2022). Traditional scoring models have suffered from significant efforts needed to collect training data and develop algorithmic models, while LLMs hold a distinct advantage in addressing this challenge. This study specifically investigates how the application of LLMs with CoT to the scoring process can ease human efforts while capturing student thinking in constructing scientific explanations, aligning more closely with human scoring outcomes. Using an experimental design, we examined LLMs’ scoring accuracy under different conditions, specifically controlling for variables of prompting approach (zero-short vs. few short learning), LLM reasoning strategy (CoT vs. Non-CoT), and provision of contextual item information and scoring rubrics (CR). We further tested the effect of versions and hyperparameters of the GPT family (i.e., ChatGPT/GPT-3.5 and GPT-4) on the automatic scoring performance. The study addresses four research questions (RQs):

RQ 1. How do GPT-3.5 and GPT-4 automatically generate explainable scores?

RQ 2. How accurate are GPT-3.5 and GPT-4 in automatically scoring student-written explanations of scientific phenomena under varying conditions (zero-shot, few-shot, CoT, CR)?

RQ 3. To what extent does CoT improve GPT-3.5 and GPT-4's automatic scoring accuracy under various conditions?

RQ 4. Which LLM models (i.e., GPT-3.5 and GPT-4) yielded better-scoring accuracy using the voting vs. single-call approach?

## 2 Literature Review

In this section, we review four strands of existing literature relevant to this study. The first section reviews previous approaches to developing automatic scoring models, before the rise of LLMs. We then review the concept and strength of LLMs that are promising for automatic scoring. In the section to follow, we examine prompt engineering strategies widely used to exploit LLMs for various purposes. At last, we summarize established methods to yield accurate results from LLMs, which are used in this study to fully elicit the GPT family’s potential for automatic scoring.

### 2.1 *Existing Approaches to Developing Automatic Scoring Models*

Automatic scoring of student-written responses to science assessment items leverages text classification NLP techniques in supervised machine learning. Previous studies have succeeded in achieving high machine-human agreement, developing automatic scoring models following the typical and labor-intensive process of machine learning: collecting student responses and scoring them, developing algorithms and training the model, and testing the model performance to further employ various strategies to improve it (Nehm et al., 2012; Wilson et al., 2023; Zhai, in press).

To construct a robust scoring model, a substantial amount of data collection is imperative. Researchers typically need to gather a significant volume of student responses, ranging from hundreds to over a thousand, to ensure a diverse and representative dataset. This process is critical for the development of an accurate and reliable model. Once these responses are compiled, trained human experts are employed to evaluate and score them. Their assessments serve a dual purpose: they not only provide a benchmark for the model's performance but also generate crucial training and testing data. This data is then used to train the model, teaching it to recognize and evaluate key elements in student responses. The involvement of human experts ensures that the model's scoring aligns with educational standards and objectives, thereby enhancing the model's utility in real-world educational settings. This rigorous process of data collection and expert evaluation forms the foundation of a normal scoring model, setting the stage for it to accurately and effectively assess student responses.

More than a decade ago, researchers started to develop automatic scoring algorithmic models based on tokens in student answers (bag of words) (Leacock & Chodorow, 2003; Ramesh & Sanampudi, 2022). After tokenizing student-written answers to items, the input data were converted to a document-term matrix or term-document matrix (in case of automatic scoring, a student's answer to an item can be considered a 'document'). Later, a document was represented as a vector with multiple dimensions, as many as the kinds of tokens used for various calculations for classification. Usually, the number of dimensions of the documents is reduced by word or document embedding, which is typically done by kernels, principal component analysis, or a neural network (Cozma et al., 2018; Birunda & Devi, 2021). After embedding, features or semantics of student-written responses to items can be processed in sophisticated neural network-based algorithms. Aligned with the data structure, recurrent neural network-based models such as long-short term memory, attention, and transformer enabled considering the context of the document, which is defined by the (bidirectional) distribution of certain tokens or characters (Haller et al., 2022). After processing student answers using these algorithms, the classification layers could yield the softmax probabilities to determine the label for a student's written response.

To further improve the scoring model performance, researchers have adopted the strategy of combining predictions from various scoring models to determine the final label. This ensemble approach has been applied to assess students’ argumentation (Wilson et al., 2023; Zhai et al., 2023), explanations (Jescovitch et al., 2019), and teachers’ pedagogical content knowledge (Zhai et al., 2020a), and showing superior results. However, this ensemble approach is part of a larger process encompassing data preparation, model development, training, and testing, which has proven to be a formidable obstacle, especially for educational researchers not versed in computer programming and machine learning methodologies.

Consequently, there is a pressing need to overcome these technical barriers in the field of automatic scoring research. By simplifying these complexities, the potential for AI-enhanced scoring innovations can be expanded, making them more accessible and beneficial to a wider range of educational professionals. This initiative is crucial in leveraging the full capabilities of generative AI such as ChatGPT in educational settings (Zhai, 2023a), ensuring that advanced scoring models are not just reserved for those with technical expertise but are also available to educators who can most directly apply these tools in their teaching and assessment practices.

Moreover, prior research on automatic scoring has frequently overlooked the explainability of the scoring outcomes (Hahn et al., 2021; Korkmaz et al., 2019). Many sophisticated scoring models, such as neural networks, encode information in an abstract mathematical space with their intricate architectural structure, which makes it difficult to understand how each factor contributes to the final model prediction (Bearman & Ajjawi, 2023). This black box issue necessitates explainability so that users may establish trustworthiness in automatic scoring, which has gained increasing recognition (Holzinger et al., 2022). This matter is especially pertinent to education, where AI models process sensitive data (i.e., student responses), and the outcomes can significantly impact teachers' instructional decision-making (Khosravi et al., 2022; Gillani et al., 2023, Muhamedyev et al., 2020; Hitron et al., 2019). The imperative for transparency becomes essential when these models are employed for formative assessment of students' responses to scientific practices (Zhai, 2021). In such scenarios, educators must understand the rationale behind the assigned grades and the criteria used to effectively provide authentic pedagogical support. Thus, enhancing the explainability of AI models in educational settings is not just a technical challenge but also a fundamental ethical consideration, ensuring that these technologies align with the educational objectives and support effective teaching practices.

### 2.2 *Large Language Model for Automatic Scoring*

LLMs such as Google pre-trained BERT (Bidirectional Encoder Representations from Transformers; Devlin et al., 2018), SciEdBERT (specialized for science education) (Liu et al., 2023), and GPT variants (Latif & Zhai, 2024) have become visionary instruments for automatic scoring in the rapidly developing field of AI. There are now new possibilities in educational assessment and other fields thanks to their unmatched capacity to process, comprehend, and generate natural language. This research program sheds light on exploring the fusion between advanced technology and practical usability, evidencing to the ever-growing capabilities of LLMs.

#### **2.2.1** *Strengths of Large Language Models in Automatic Scoring*

LLMs have shown significant advancements in educational assessment and automatic scoring. A pivotal moment in this field was marked by the introduction of BERT by Devlin et al. (2018). BERT's deep bidirectional training fundamentally enhanced the understanding of language context, a critical factor in the effectiveness of scoring applications where nuanced interpretation of text is essential. Building upon this foundation, recent studies such as Lee, Jung et al. (2023) have highlighted the efficiency of models like GPT-3.5 in few-shot learning scenarios. This approach, requiring minimal examples to generate or score content effectively, is particularly valuable in educational settings characterized by varied and complex responses.

Further emphasizing the versatility of LLMs, research by Organisciak et al. (2023) has demonstrated their potential in scoring tasks involving divergent thinking. While traditionally challenging for automated systems, this area has seen significant improvement with LLMs that are now capable of assessing creativity and originality beyond mere semantic analysis. In the realm of automated essay scoring, the work by Rodriguez et al. (2019) reveals that LLMs offer higher accuracy and reliability, addressing many of the limitations of earlier scoring systems. This advancement is crucial in ensuring fair and comprehensive evaluation of complex written responses. The potential of LLMs, particularly GPT-3.5, in educational applications extends beyond scoring. They highlight how generative AI can significantly support and enhance teaching and learning processes, making these advanced technologies accessible and beneficial for educators and learners (Baidoo-Anu & Ansah, 2023). LLMs' potential goes beyond simple scoring of student responses - rather, it streamlines teachers' assessment practices and facilitates providing feedback to students. A remarkable work by Bewersdorff et al. (2023) provided a foundation for productive and personalized feedback and found that GPT-4 can accurately identify errors in student response than human rater. These works also highlight the potential use of LLMs for education, specifically when fine-tuning models like GPT-3.5 using challenging mathematical datasets (Latif & Zhai, 2023) can provide more insights about its application in education.

In summary, the strengths of LLMs in automatic scoring are deemed multifaceted. They exhibit an advanced understanding of language, demonstrate efficiency in adaptive learning scenarios, accurately score complex cognitive tasks, and hold the potential to transform educational practices through their user-friendly AI capabilities.

#### **2.2.2** *Current Research Trends in LLMs for Automatic Scoring*

The landscape of automatic scoring is being reshaped by the advent of LLMs, with recent studies revealing diverse applications and exploring their potential limitations. Expanding on the foundation of BERT, several studies have explored the application of LLMs in specific educational settings. For instance, researchers have focused on pre-training strategies tailored for science education, demonstrating how context-specific adaptations can enhance the effectiveness of LLMs in scoring and content generation for science-related tasks (SciEdBERT - Z. Liu et al., 2023; Wu et al., 2023). Similarly, another study (Shen et al., 2021) introduced MathBERT, a model specifically designed for mathematics education, showcasing the potential of subject-specialized LLMs in automatic scoring.

The emergence of GPT, particularly in its GPT-3.5 variant, has introduced a new paradigm in using LLMs for educational purposes (Zhai, 2023b). Studies also highlight GPT3.5's effectiveness in few-shot learning scenarios and its application in educational content generation. These works underline the model's efficiency in understanding and generating nuanced language with minimal input, surpassing the capabilities of earlier LLMs like BERT in certain aspects, particularly in user-friendly interaction and adaptability to diverse educational needs (Lee, Jung et al., 2023; Rahman & Watanobe, 2023).

However, the deployment of LLMs in education is not without challenges. Studies also provide critical perspectives on the implications of LLMs like GPT-3.5 in traditional assessments and the ethical considerations in educational settings. Research points to the need for careful evaluation of the impact of these models on traditional learning and assessment methodologies, highlighting the importance of addressing potential biases and ethical dilemmas, which can significantly impact teachers' assessment practices and thus provide customized support to students (Rudolph et al.,2023; Yan et al., 2023).

To sum up, current research trends in LLMs for automatic scoring are increasingly focusing on the interplay between sophisticated general language understanding and the specific needs in educational applications. Specialized models like SciEdBERT have emerged, building upon the BERT architecture and targeting specific educational domains (Z. Liu et al., 2023). Studies also explored context-specific pre-training strategies, indicating the importance of tailoring LLMs to particular educational subjects for improved performance in automatic scoring (Z. Liu et al., 2023; Wu et al., 2023). This study builds upon these trends by exploring the practical implications of using advanced LLMs in a novel educational context, emphasizing the need for balanced and ethical deployment of these technologies in educational settings.

### 2.3 *Prompt Engineering with Chain-of-Thought*

Prompt engineering (P. Liu et al., 2023) focuses on enhancing the in-context learning ability of LLMs by designing more efficient prompt templates. However, utilizing LLMs for complex reasoning tasks remains a significant challenge, even with refined prompt engineering techniques. For example, LLMs sometimes fail to solve logical reasoning or arithmetic problems when they are presented in intricate statements (Jung et al., 2022; Zhou et al., 2022). Given this, it is crucial to recognize that automatic scoring is a complex reasoning task, as there is a strict rubric to guide the grading process. Especially in science education, the levels of scores are usually designed as the essential steps of a reasoning path, from the known to the answer (Zhai, He, & Krajcik, 2022).

Recently, Wei et al. (2022) found that guiding LLMs to perform the task following a reasoning path could significantly improve their capability in solving such challenging tasks, known as chain-of-thought (CoT) prompting. CoT is defined as "a series of intermediate natural language reasoning steps that lead to the final output" (Wei et al., 2022) (p. 2). The most straightforward way to achieve this goal is to encourage LLMs to generate their reasoning paths before making the final prediction, which could be easily made by using a magic prompt, i.e., "Let's think step by step", as suggested by Kojima et al. (2022). With this prompt provided, the LLMs would provide the rationales for the given tasks and follow the self-generated rationales to make the final prediction, called the zero-shot CoT (Kojima et al., 2022). Following this path, many advanced strategies are proposed to improve the rationales generated by the models (Besta et al., 2023; Wang et al., 2022; Yao et al., 2023), and they are widely adapted to diverse tasks and various scenarios, such as programming (Bi et al., 2023; Cheng et al., 2022), math problems (Imani et al., 2023), multi-modalities question answering (Chen et al., 2023).

The self-generated CoT approach has shown potential in utilizing LLMs for various reasoning tasks. However, it may not be ideally suited for specific tasks like automatic scoring. This limitation arises because grading rubrics, especially those crafted by educators for individual items, often follow unique reasoning paths not typically encountered by LLMs during their pre-training phase. Consequently, LLMs might struggle to generate valid reasoning paths aligned with these specialized grading processes. To address this challenge, our study tries the few-shot CoT method (Wei et al., 2022) for automatic scoring. In this approach, the prompt template will incorporate a selection of student responses, each accompanied by a CoT demonstration guided grading score written by human graders.

To sum up, we compare the results of zero-shot and few-shot CoT prompt engineering of LLM on grading student-written answers, which of both have rarely been tried in automatic scoring research. This setup aims to help the LLMs follow the reasoning paths behind the example CoT demonstrations, thereby enhancing their predictive accuracy in the context of grading.

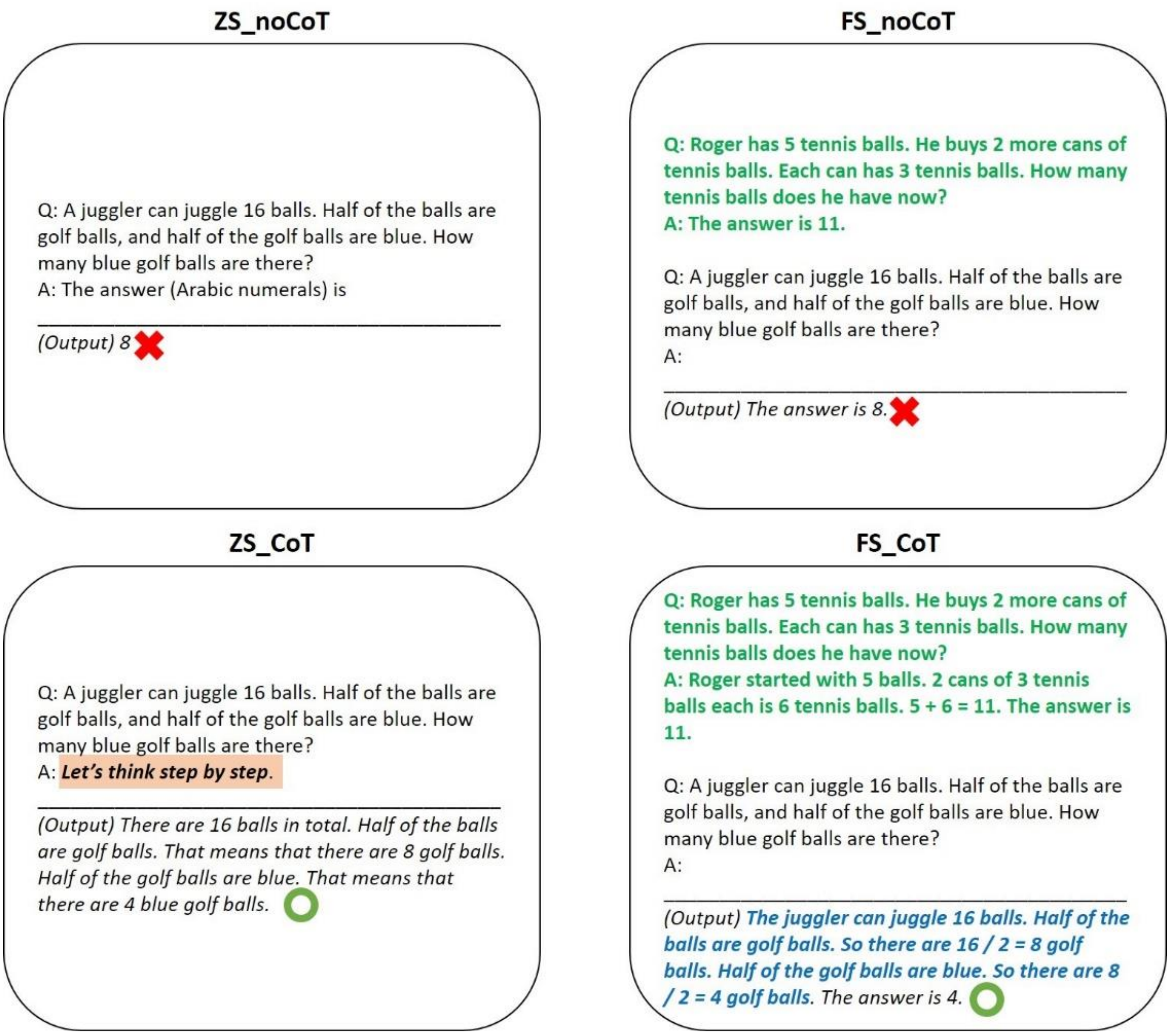

**Fig. 1** Typical Examples of Zero-Shot and Few-Shot Learning Prompt Engineering (ZS: Zero-Shot, FS: Few-Shot, CoT: Chain-of-Thought) (reorganized from Wei et al., 2022 and Kojima et al., 2022)

### 2.4 *Yielding Reliable Results from GPT*

Although it is promising to apply LLMs for automatic scoring with the zero/few-shot CoT strategy, it faces a significant challenge due to the inherent uncertainty in LLMs' generative process. This uncertainty is introduced by the sampling strategies used to produce diverse responses (Hewitt et al., 2022; Holtzman et al., 2019; Li et al., 2022; Meister et al., 2023; Su et al., 2022). LLMs predict the likelihood of each word in a sequence, aiming to create responses with the highest joint probability over each word of the sequences. However, considering the vast number of potential combinations on the order of $O(V^N)$, where $V$ is the number of candidate words and $N$ is the response length, it is impractical to evaluate all possible responses.

The Greedy Sampling is the most naive solution for this problem, which generates the word with the maximum likelihood at each step. If the problem space being searched by the model has a characteristic that the minima found by the greedy sampling are the global minima, not a local minimum, greedy sampling can be the most effective and efficient way to solve the problem. Practically, setting hyperparameters as temperature = 0.0 and top_p = 0.01 for GPT is used to take the greedy sampling approach.[1] However, it could lead to poor performance in generating a long text without a powerful enough LLM (e.g., GPT-4 has more parameters than GPT-3.5) since the estimations of word likelihood can be incorrect (Fu et al., 2021; Holtzman et al., 2019).

Another common approach is Nucleus Sampling (Holtzman et al., 2019). By allowing a dynamic nucleus of the probability distribution, the quality of machine generated responses could be increased. Practically, setting hyperparameters as temperature = 0.9 and top $p$ = 0.95 for GPT is used to take the nucleus sampling approach. However, this approach could lead to different outputs over the same input, introducing uncertainty to the grading process.

To overcome this issue, we propose to append a voting strategy to aggregate the prediction scores generated by multiple calls. Specifically, we could call GPT API multiple times and consider the most frequent predicted label mentioned by the responses as the final prediction. In this way, we increase the reliability of the advanced nucleus sampling. This approach is inspired by both strands of research: (1) machine learning fields ensembling the answers from multiple models, and (2) educational studies asking two or three educators to independently grade the same student submission. By incorporating multiple grading instances, we aim to decrease the variance of the predictions, thereby enhancing the reliability and consistency of the grading process.

[1] Temperature can span 0-2 and higher values make the output more random and lower values make it more focused and deterministic; Also, top_p designates the probability mass the model considers - e.g., top_p = 0.1 means only the tokens comprising the top 10% probability will be considered in text generation. For details, see https://platform.openai.com/docs/api-reference/chat/create

The issue of yielding reliable results from LLMs is a complicated matter since it is related to the power of the model itself. For example, using greedy sampling or nucleus sampling with GPT-4 could return different results from using these with GPT-3.5, since the former has broader general knowledge and problem solving abilities (OpenAI, 2023). However, the impact of these mixed conditions on the performance of LLMs, particularly of GPT, on automatic scoring has not been explored yet, to our knowledge.

In this study, we conducted experiments using both greedy sampling and nucleus sampling, with both LLMs that are more powerful (GPT-4) or less powerful (GPT3.5). Our study distinctively compares the performance of GPT-4 and GPT-3.5 in handling complex, authentic student responses to science items. This comparative aspect is critical in highlighting the evolutionary strides in LLMs and their practical efficacy in educational contexts. Our unique contribution lies in our methodological approach: we analyzed extensive data sets, prioritize the authenticity of student responses, and focus on obtaining explainable and interpretable results. This approach not only benchmarks the performance differences between GPT-4 and GPT-3.5 but also sets a new standard in evaluating the practical utility of LLMs in educational assessments -- comparing different prompt engineering methods and hyperparameters simultaneously. Our findings offer novel insights into the operational dynamics of these models, contributing significantly to the growing body of knowledge on the effective use of LLMs in educational settings. The details of the experiments and the results are presented in Methods and Findings.

# 3 Methods

## 3.1 *Dataset*

This study conducted a secondary analysis of a dataset that asked middle school students in U.S. to describe scientific models accounting for science phenomena (details see Zhai, He, & Krajcik (2022)). Specifically, we employed six assessment tasks, three with binomial scoring rubrics (Tasks R1_2, J2_2, and J6_2) and three with trinomial scoring rubrics (Tasks H4_2, H4_3, and J6_3). The tasks were designed to examine whether students meet the NGSS performance expectation, MS-PS1-4 (MS: Middle School, PS: Physical Sciences). *Develop a model that predicts and describes changes in particle motion, temperature, and state of a pure substance when thermal energy is added or removed*. For each task, Zhai, He and Krajcik (2022) collected more than 1,000 student responses. All the items were scored using corresponding rubrics to examine students' proficiency levels (i.e., Beginning, Developing, and Proficient), and the inter-rater reliability for each item was over Cohen's kappa = 0.75. We used the dataset from the parental study with ground-truth labels given by human scorers.

To examine the scoring accuracy of LLMs, we randomly sampled student responses from the existing dataset to make a balanced testing dataset for each task to avoid unnecessary errors, potential bias in model fitting and possible inflation/detriment in performance (Fang et al., 2023). For example, since the original dataset was dominated by 'Beginning' cases (~ 78% in the original task H4_2 data), even if a model just predicts every test case into 'Beginning' without exception, it could have shown seemingly but delusively high accuracy. Consequently, we randomly selected 120 student responses ranked at 'Beginning' for all the tasks and randomly selected cases ranked at 'Proficient' and 'Developing' as close to 120 as possible, respectively, depending on the available responses at the respective levels. We found that besides Task H4_2 with 110 'Proficient' cases and 80 'Developing' cases, and J6_3 with only 20 'Proficient' cases, each of the other tasks provided 120 testing cases at each proficiency level. This sampling approach resulted in 1,650 student-written responses in the test dataset (see 1).

**Table 1** Task IDs and Number of Cases (*N* = 1,650)

| Task | Total | 'Proficient' | 'Developing' | 'Beginning' |
|---|---|---|---|---|
| R1_2 | 240 | 120 | NA | 120 |
| J2_2 | 240 | 120 | NA | 120 |
| H4_2 | 310 | 110 | 80 | 120 |
| H4_3 | 360 | 120 | 120 | 120 |
| J6_2 | 240 | 120 | NA | 120 |
| J6_3 | 260 | 20 | 120 | 120 |

## 3.2 *Experimental Design*

We conducted experiments using various combinations of prompt engineering approaches to compare the performance of those in automatic scoring.

The first aspect we tested was zero-shot and few-shot learning. For zero-shot learning, we did not provide GPT-4 with any example of human coders' evaluation of student written responses. In contrast, we provided GPT-4 with four examples of human coders' evaluation for few-shot learning. Note that these four few-shot cases were not included in the test dataset. The second aspect we tested was the use of CoT, which formulated three conditions: prompts without CoT, with CoT, and with CoT plus scoring rubric and problem context. Consequently, we tried six prompts to automatically score each item. The details of prompts are presented in Prompt Engineering.

We mainly used GPT-4 API with hyperparameters of temperature = 0 and top_p = 0.01 for automatic scoring, which is expected to give the most reliable results by greedy decoding. We conducted additional experiments to compare our approach's automatic scoring performance with others. We adopted three additional approaches that use GPT-4 or GPT-3.5.[2] First and second approaches call GPT-4 or GPT-3.5 API thrice with hyperparameters of temperature = 0.9 and top_p = 0.95 and holds a vote to determine the label of the test case. The labels of almost every test case could be decided through this process (e.g., if there are two for 'Developing' and one for 'Beginning,' the label is determined as 'Developing'), while 12 and 15 cases among 930 trinomial classification cases had no majority prediction (i.e., one for 'Proficient,' one for 'Developing,' and another one for 'Beginning'). We called GPT-4 or GPT-3.5 API once again to determine the label for those cases. The third approach, calls GPT-3.5 once with hyperparameters of temperature = 0 and top_p = 0.01.

Note that all four approaches (calling GPT-4 once, GPT-4 thrice, GPT-3.5 once, and GPT-3.5 thrice) could be considered as ways to receive reliable classification results from the GPT model family, as explained in the Literature Review. The summary of the experimental design is presented in Table 2.

[2] We used gpt-4 and gpt-3.5-turbo models in OpenAI API calls throughout this study. See https://platform.openai.com/docs/models/overview

**Table 2** Experimental Design (above: acronyms of used prompts, below: large language model settings

| Types of Prompt used | | | |
|---|---|---|---|
| | No Chain-of-Thought | Chain-of-Thought | Chain-of-Thought with Context and Rubric |
| Zero-shot | ZS_noCoT | ZS_CoT | ZS_CoT_CR |
| Few-shot | FS_noCoT | FS_CoT | FS_CoT_CR |

| Large Language Model Used | | | | |
|---|---|---|---|---|
| Sampling Strategy | Greedy | Nucleus | Greedy | Nucleus |
| GPT Version | 4 | 4 | 3.5 | 3.5 |
| Hyperparameters (temperature, top_p) | (0, 0.01) | (0.9, 0.95) | (0, 0.01) | (0.9, 0.95) |
| Number of API Call | 1 | 3 | 1 | 3 |

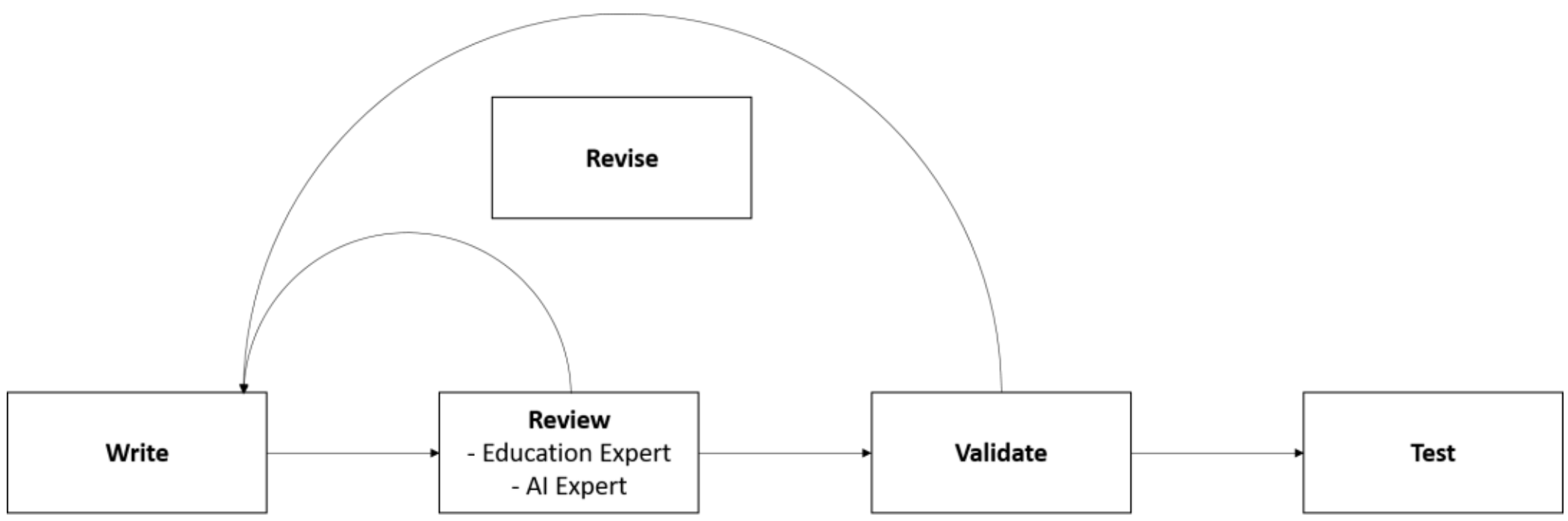

**Fig. 2** Iterative Procedure of Prompt Engineering for Educational Studies (WEVRT)

### 3.3 *Prompt Engineering*

This study proposed an iterative procedure of prompt engineering for educational studies, including four major components—prompt writing, reviewing, validating, revising, and testing (WRVRT). First, researchers write a prompt for the automatic scoring of the given item. Second, more than one expert from educational studies (in this case, from educational assessment) and AI, respectively, review the prompt. In this stage, the internal validity of the prompt is secured by means of face validity. If needs for modification arise, researchers revise the prompt. Next, researchers validate the prompts by automatic scoring of student responses that are not included in the test data. By using the validation cases, external validity of the prompt is secured. If needs for modification arise in terms of model performance, researchers will revise the prompt again. The iterative WRVRT is completed until the prompts reach saturation. Finally, researchers can run the test cases with the prompts to examine the scoring accuracy 2.

Using WRVRT, the first author with expertise in chemistry education and automatic scoring research wrote the initial prompts. Another researcher with expertise in K12 science education and automatic scoring research, one expert in large language models and machine learning, and one doctoral student in computer science reviewed the prompt. The need to revise the prompts was pointed out twice in the review stage and once in the validation stage. We finalized our prompts after three revisions.

Our prompt engineering combined six components to generate six types of prompts for each task (Table 2).

- *BasicRole* instructs GPT's role as an evaluator of student responses. *BasicRole* is transferred as the 'role' of 'system' in the GPT API call. for the prompts that provide GPT with *ContRubTEXT*, a sentence that instructs GPT to refer to *ContRubTEXT* is concatenated with *BasicRole*.
- *ContRubTEXT* first describes the stem of the assessment item, which was provided to students when they addressed the item. And then, it describes the scoring rubric for each item. Each scoring rubric lists 2-4 scoring components and, based on which, the holistic categories of 'Proficient,' 'Developing,' and 'Beginning' are determined.
- *FewEXAMPLES* provides four student-written responses with human scores for the three proficiency levels– 'Proficient,' 'Developing,' and 'Beginning.'
- *CoT Initiator* instructs GPT to develop its reasoning according to CoT. For zero-shot learning, "Let's think step by step" serves this purpose. For few-shot learning, the prompt component provides four human scoring examples with human evaluator-written CoT, as well as the category of 'Proficient,' 'Developing,' and 'Beginning.' For example, the human evaluator exemplifies which part of the student-written answer can be considered evidence of each component in the scoring rubrics or point out that there is no evidence of it.

After investigating the answer for all the components, the human evaluator synthesizes it to decide the holistic category.

Table 3 shows the inclusion of each component according to the types of prompts. Note that for zero-shot with CoT prompt or zero-shot with CoT with problem context and rubric, "Let's think step by step" was concatenated at the end of the prompt to provoke CoT reasoning of GPT. Also, Figure 3 compares the six prompt engineering strategies. The comprehensive examples of the six components are presented in Appendix 1.

### 3.4 *Data Analysis*

We conducted the experiment on Python 3.10 environment, with GPT-4 and GPT3.5-turbo APIs provided by OpenAI. After collecting GPT family APIs' classification of student-written responses, accuracy, precision, recall, and F1 were calculated by comparing GPT labeling with human consent labeling.

**Table 3** Combinations of Prompt Components According to the Prompt Engineering Type

| | BasicRole | ContRubTEXT | FewEXAMPLES | CoT Initiator |
|---|---|---|---|---|
| ZS_noCoT | Y | N | N | N |
| ZS_CoT | Y | N | N | "Let's think step by step" |
| ZS_CoT_CR | Y | Y | N | "Let's think step by step" |
| FS_noCoT | Y | N | Y | N |
| FS_CoT | Y | N | Y | Examples of human scoring |
| FS_CoT_CR | Y | Y | Y | Examples of human scoring |

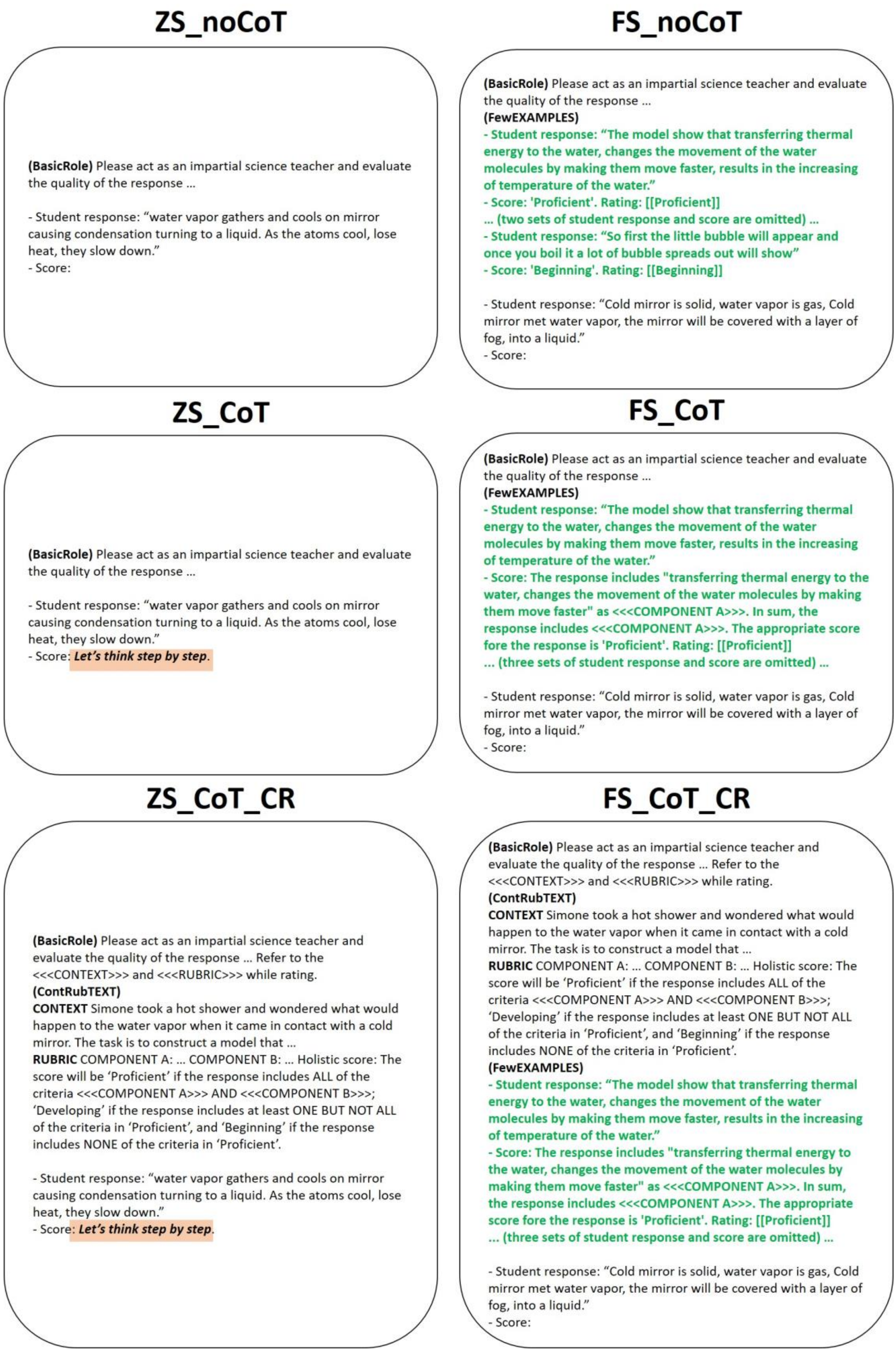

**Fig. 3** Examples of the Six Prompts Used for Automatic Scoring (Task H4_3; ZS: Zero-Shot, FS: Few-Shot, CoT: Chain-of-Thought, CR: Problem Context and Rubric)

# 4 Findings

In this section, we first exemplify how GPT-4 responds to the automatic scoring query from the users (RQ 1). We then present the scoring accuracy of various strategies to answer RQs 2-3. At last, we present the comparison of performance between different GPT versions according to hyperparameters compared to answer RQ 4.

## 4.1 *Responses of GPT to Automatic Scoring Queries*

The results show that LLMs can respond to automatic scoring tasks, providing the user with explainable responses. Figure 4 shows the examples of GPT-4's responses to the automatic scoring task (H4_3) (see Figure 3 for the prompt components). Specifically, in every prompt, GPT-4 returned the reason for why it classified a student's response into a specific category - 'Proficient,' 'Developing,' or 'Beginning,' other than FS_noCoT. Exceptionally, FS_noCoT prompt made GPT-4 return simple classification results, following the example of human grading provided in *FewEXAMPLES*.

We found that zero-shot prompts returned relatively longer responses, allowing GPT4 to spontaneously generate reasons for its classification. In contrast, few-shot prompts returned relatively shorter responses, strictly following the structure given in the few-shot examples. Especially, FS_CoT and FS_CoT_CR generated answers according to the CoT structure developed in this study for automatic scoring. Note that in the examples in Figure 4, ZS_CoT_CR, FS_CoT, and FS_CoT_CR made correct predictions.

## 4.2 *Scoring Accuracy of LLMs by Prompting Strategy*

The scoring accuracy of various prompting strategies is presented in Table 4. There was no single prompting that showed the best accuracy for all tasks; instead, the best-performing prompting differs by the item, though showing some patterns. Overall, prompt engineering works better for binomial items as compared with trinomial items.

Specifically, the accuracy was found to be up to .9083 (J6_3), .8792 (J2_2), and .7833 (R1_2) for the binomial items and up to .6806 (H4_3), .5935 (H4_2), and .5885 (J6_3) for the trinomial items. Other metrics, such as Precision, Recall, F1, and Quadratic Weighted Kappa for each item according to prompting, are presented in Appendix 2. Below, we uncovered some patterns according to the experiment.

**Table 4** Test Accuracy (Standard Deviation) of GPT-4 for the Items by Prompt Engineering Strategies (ZS: Zero-Shot, FS: Few-Shot, CoT: Chain-of-Thought, CR: Problem Context and Rubric; **Bold**: Best Accuracy for the Item Among the Prompt Engineering Strategies)

| Item | Type | ZS_noCoT | ZS_CoT | ZS_CoT_CR | FS_noCoT | FS_CoT | FS_CoT_CR |
|---|---|---|---|---|---|---|---|
| R1_2 | Binomial | 0.6625 | 0.6458 | 0.7583 | **0.7833** | 0.7500 | 0.7625 |
| J2_2 | Binomial | 0.6417 | 0.6417 | 0.8458 | 0.7958 | 0.8375 | **0.8792** |
| H4_2 | Trinomial | 0.3613 | 0.3710 | **0.5935** | 0.5581 | 0.5774 | 0.5452 |
| H4_3 | Trinomial | 0.4722 | 0.5111 | 0.6333 | 0.5917 | **0.6806** | 0.6667 |
| J6_2 | Binomial | 0.6583 | 0.6458 | 0.6792 | 0.7833 | 0.8250 | **0.9083** |
| J6_3 | Trinomial | 0.4962 | 0.5038 | **0.5885** | 0.4500 | 0.2385 | 0.4231 |
| Average | | 0.5487 (0.1135) | 0.5532 (0.102) | 0.6831 (0.0927) | 0.6604 (0.1342) | 0.6515 (0.2047) | 0.6975 (0.1737) |
| | | 0.595 (0.1205) | | | 0.6698 (0.1744) | | |

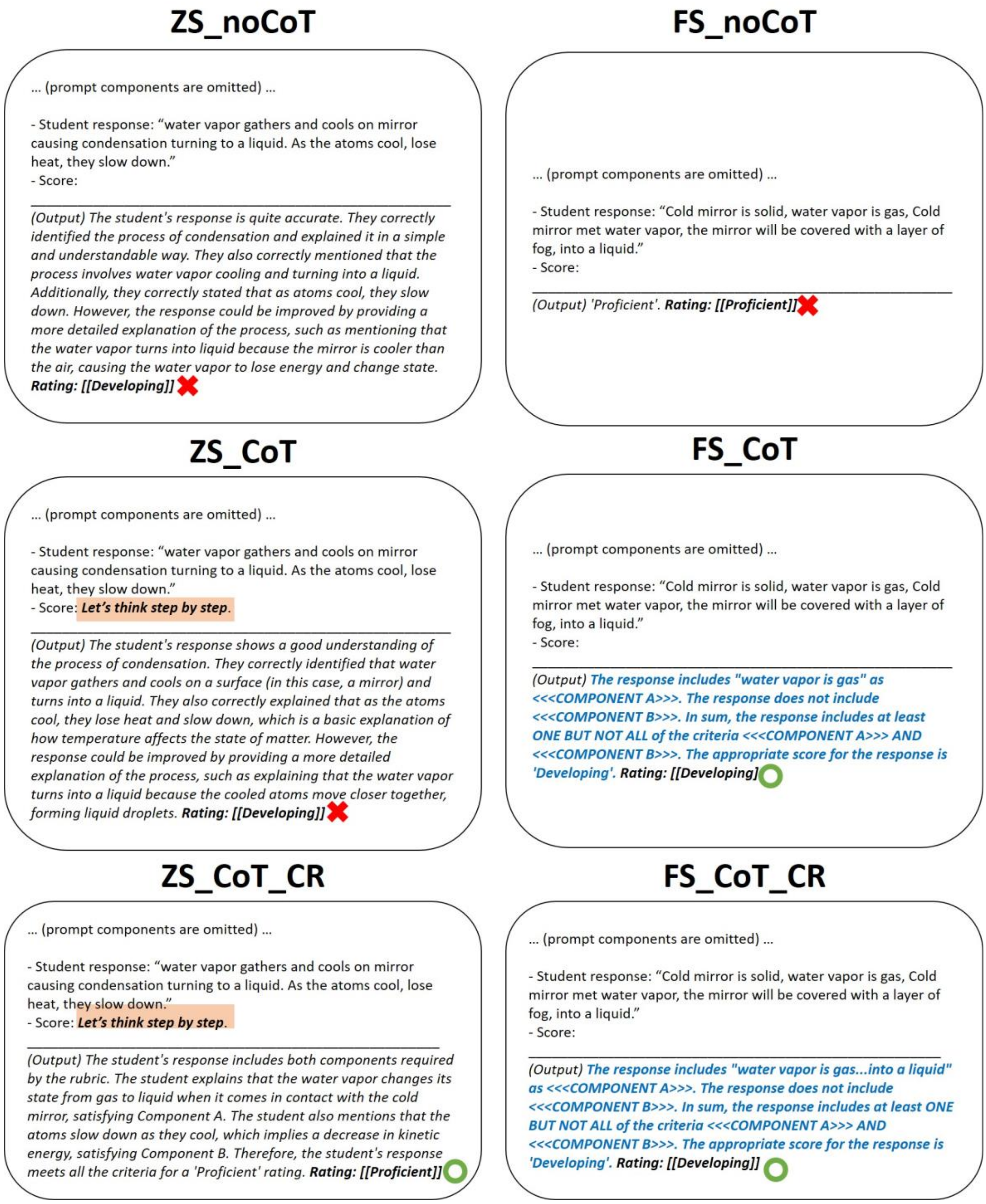

**Fig. 4** Example Responses of GPT According to Prompting Strategy (Task H4_3; ZS: Zero-Shot, FS: Few-Shot, CoT: Chain-of-Thought, CR: Problem Context and Rubric)

### 4.2.1 *Zero-Shot vs. Few-Shot Learning Prompts*

We first compared the scoring accuracy between zero-shot and few-shot learning. On average, we found that few-shot learning showed a higher average scoring accuracy (M = .6698; SD = .1744) compared to zero-shot learning (M = .595; SD = .1205), with an increase of 12.6%. Specifically, our testing suggests that zero-shot (ZS_noCoT) yielded an average scoring accuracy, M = .5487 (SD = .1135), while few-short learning (FS_noCoT) demonstrated a higher average scoring accuracy, M = .6604 (SD = .1342) with an average increase of 20.4% throughout six items. These results serve as the baseline of the concurrent GPT family's performance on automatic scoring of student-written responses, indicating that few-shot learning significantly outperformed zero-shot learning on automatic scoring.

We found a similar pattern between zero-shot and few-shot learning with the CoT strategy. Specifically, our results show an average increase of 17.8% in scoring accuracy from zero-shot with CoT (ZS_CoT; M = .5532, SD = .102) to few-shot with CoT strategies (FS_CoT; M = .6515, SD = .2047). Interestingly, we found a decreased difference (2.1%) in average scoring accuracy from few-shot (FS_CoT_CR; M = .6975, SD = .1737) to zero-shot (ZS_CoT_CR; M = .6831, SD = .0927) learning when using both CoT and CR.

#### **4.2.2** *Effects of Chain-of-Thought Prompt Engineering*

To examine how CoT impacts automatic scoring accuracy (RQ 3), we first checked the best scoring performance strategies for each task and then compared the average performance with and without CoT. We found that five out of the six tasks scored best with CoT, except for one task (R1_2). Specifically, H4_3 showed its highest accuracy = .6806 with FS_CoT, H4_2 (acc = .5935) and J6_3 (acc = .5885) with ZS_CoT_CR, and J2_2 (acc = .8792) and J6_2 (acc = .9083) with FS_CoT_CR. These results suggest that prompts with CoT yielded the best scoring performance for most tasks.

We found that CoT was especially useful when pairing with CR. Among the five highest-performing prompts mentioned above, four showed the highest scoring accuracy with CoT_CR. Similar evidence was found in zero-shot and few-shot learning, respectively. ZS_CoT_CR yielded a scoring accuracy = .6831, which is about 13.44% higher than ZS_noCoT on average, while few-shot learning with CR (FS_CoT_CR) showed an average scoring accuracy = .6975, which is 3.7% higher than FS_noCoT on average.

In contrast, we found that CoT made a limited contribution to scoring accuracy without pairing with CR. Specifically, ZS_noCoT showed average accuracy = .5487 throughout the six items, and ZS_CoT showed average accuracy = .5532, which is only a .82% improvement. Likewise, FS_noCoT showed an average accuracy = .6604, and FS_CoT showed an average accuracy = .6515, a slight decrease (-1.35%). The less than a 2% difference brought by both cases suggests that CoT without CR contributed limited to the automatic scoring performance of GPT-4.

#### **4.2.3** *How Chain-of-Thought Prompt Engineering Functions*

To uncover how CoT functions with CR, we investigated its performance by digging into category-wise test accuracy of scoring prompts (Table 5). CoT_CR seems to increase the overall accuracy by balancing accuracy for all proficiency categories ('Proficient,' 'Developing,' or 'Beginning'). For example, in task J2_2, ZS_CoT_CR (acc = .8458) increased the scoring accuracy of ZS_CoT (acc = .6417) by 31.81% (Table 4). This change diverged at the two proficiency levels. While scoring accuracy decreased by .94% for the 'Proficient' category from .8833 to .875, it changed from .4 to .8167 for the 'Beginning' category, improving by 104.2% (Table 5). Likewise, from FS_CoT to FS_CoT_CR, average scoring accuracy for the 'Proficient' level increased by 20.0% while that for the 'Beginning' level decreased by 6.0% (Table 5). Consequently, FS_CoT_CR (acc = .8792) showed 4.98% higher accuracy than FS_CoT (acc = .8375) (Table 4).

The category-wise balancing effect of CoT_CR is more obvious in trinomial tasks. For example, in task H4_2, ZS_CoT_CR (acc = .5935) increased the scoring accuracy of ZS_CoT (acc = .3710) by 59.97% (Table 4). This change diverged at the three proficiency levels. While scoring accuracy decreased by 9.84% for the 'Developing' category from .7625 to .6875, it changed from .275 to .6833 for the 'Beginning' category, improving by 148.47%, and from .1909 to .4273 for the 'Proficient' category, improving by 123.83%. Consequently, the Quadratic Weighted Kappa changed from .2525 (ZS_CoT) to .5806 (ZS_CoT_CR), improving by 129.94%. This increase of Quadratic Weighted Kappa accompanied with balanced category-wise accuracy was observed in every comparison of ZS/FS_CoT versus ZS/FS_CoT_CR in tasks H4_2, H4_3, and J6_3, except for one case (from FS_CoT to FS_CoT_CR in task H4_2). Convincingly, similar pattern balancing effect was found throughout the six items, with few exceptions.

**Table 5** Category-wise Test Accuracy of GPT-4 According to Item and Prompt Engineering (ZS:Zero-Shot, FS: Few-Shot, CoT: Chain-of-Thought, CR: Problem Context and Rubric)

| Task | Parameter | ZS_noCoT | ZS_CoT | ZS_CoT_CR | FS_noCoT | FS_CoT | FS_CoT_CR |
|---|---|---|---|---|---|---|---|
| R1_2 | Acc Prof | 0.7917 | 0.8333 | 0.9083 | 0.675 | 0.5167 | 0.55 |
| | Acc Dev | NA | NA | NA | NA | NA | NA |
| | Acc Beg | 0.5333 | 0.4583 | 0.6083 | 0.8917 | 0.9833 | 0.975 |
| | Kappa | NA | NA | NA | NA | NA | NA |
| J2_2 | Acc Prof | 0.8583 | 0.8833 | 0.875 | 0.7167 | 0.7083 | 0.85 |
| | Acc Dev | NA | NA | NA | NA | NA | NA |
| | Acc Beg | 0.425 | 0.4 | 0.8167 | 0.875 | 0.9667 | 0.9083 |
| | Kappa | NA | NA | NA | NA | NA | NA |
| H4_2 | Acc Prof | 0.2583 | 0.275 | 0.6833 | 0.6 | 0.475 | 0.4833 |
| | Acc Dev | 0.725 | 0.7625 | 0.6875 | 0.3625 | 0.5 | 0.8125 |
| | Acc Beg | 0.2091 | 0.1909 | 0.4273 | 0.6545 | 0.7455 | 0.4182 |
| | Kappa | 0.2791 | 0.2525 | 0.5806 | 0.4701 | 0.5583 | 0.5004 |
| H4_3 | Acc Prof | 0.5417 | 0.6083 | 0.775 | 0.7083 | 0.5667 | 0.6833 |
| | Acc Dev | 0.7417 | 0.7667 | 0.7083 | 0.225 | 0.6833 | 0.675 |
| | Acc Beg | 0.1333 | 0.1583 | 0.4167 | 0.8417 | 0.7917 | 0.6417 |
| | Kappa | 0.4258 | 0.4835 | 0.6276 | 0.6111 | 0.6667 | 0.6831 |
| J6_2 | Acc Prof | 0.9417 | 0.9667 | 1 | 0.7167 | 0.7417 | 0.8833 |
| | Acc Dev | NA | NA | NA | NA | NA | NA |
| | Acc Beg | 0.375 | 0.325 | 0.3583 | 0.85 | 0.9083 | 0.9333 |
| | Kappa | NA | NA | NA | NA | NA | NA |
| J6_3 | Acc Prof | 0.2583 | 0.275 | 1 | 0.5333 | 0.3417 | 0.75 |
| | Acc Dev | 0.8 | 0.8 | 0.2583 | 0.3417 | 0.025 | 0.05 |
| | Acc Beg | 0.1 | 0.1 | 0.1 | 0.6 | 0.9 | 0.7 |
| | Kappa | 0.186 | 0.1824 | 0.3467 | 0.3309 | .02351 | 0.4045 |

To sum up, the results (Tables 4-5) present that (1) few-shot learning prompts show about 7.48% higher accuracy than zero-shot learning, (2) addition of mere CoT to prompts does not help increasing scoring accuracy in general, and (3) CoT given with Context and Rubric improves the scoring accuracy up to 13.44% (zero-shot) or 3.71% (few-shot). If we consider ZS_noCoT as the very baseline of automatic scoring using GPT-4 with greedy sampling, FS_CoT_CR increases the scoring accuracy by 14.88%.

### 4.3 *Performances of GPT-4 vs. GPT-3.5 According to Hyperparameters*

The scoring accuracy of GPT-4 and GPT-3.5 according to hyperparameters is presented in Table 6.

To answer RQ 4, we compared the average scoring accuracy of GPT-4 and GPT-3.5 with FS_CoT_CR prompt, which yielded the best performance in Table 4. We found that automatic scoring using GPT-4 yielded higher accuracy than using GPT-3.5 in general. Specifically, in greedy sampling (temperature = 0 and top_p = 0.01), GPT-4 showed accuracy = .6975 which is higher than GPT-3.5 (accuracy = .6331) by 10.2%. In nucleus sampling (temperature = 0.9 and top_p = 0.95), GPT-4 showed accuracy = .6802, which is above GPT-3.5 (accuracy = .635) by 4.52%. Therefore, GPT-4 showed approximately 8.64% better performance than GPT-3.5 in overall sense.

However, there was no clear pattern regarding the voting strategy. As decreasing order, with GPT-4, a single-call with the greedy sampling showed better performance (acc = .6975) than the voting strategy with the nucleus sampling (acc = .6802). Specifically, only three tasks (R1_2, J2_2, and H4_3) showed higher accuracy with the single-call than the voting strategy. In contrast, for GPT-3.5, the voting strategy showed a better performance (acc = .635) than single-call (acc = .6331). That is, we found only four tasks (R1_2, J2_2, H4_3, and H4 4) showed higher accuracy with the voting strategy than the single-call. These results show the complex interaction between the model capacity and the voting strategy with different hyperparameters.

**Table 6** Test Accuracy of GPT-3.5-based Prompt Engineering (*: Calling GPT API once with temperature = 0 and top_p = 0.01, **: Calling GPT API thrice with temperature = 0.9 and top_p = 0.95, **Bold**: Best accuracy within the item)

| Item | Type | FS_CoT_CR 4_1* | FS_CoT_CR 4_3** | FS_CoT_CR 3.5_1* | FS_CoT_CR 3.5_3** |
|---|---|---|---|---|---|
| R1_2 | Binomial | **0.7625** | 0.7458 | 0.6833 | 0.6875 |
| J2_2 | Binomial | **0.8792** | 0.7625 | 0.7542 | 0.7625 |
| H4_2 | Trinomial | 0.5452 | **0.5645** | 0.5484 | 0.5548 |
| H4_3 | Trinomial | **0.6667** | 0.6528 | 0.5638 | 0.5722 |
| J6_2 | Binomial | 0.9083 | **0.925** | 0.7833 | 0.7792 |
| J6_3 | Trinomial | 0.4231 | 0.4308 | **0.4654** | 0.4538 |
| Average | | 0.6975 (0.1737) | 0.6802 (0.1717) | 0.6331 | 0.635 (0.1289) |

To investigate the divergence of GPT-4 and GPT-3.5, we dug into the responses to our prompts and found that they used the scoring criteria differently. 5 presents examples from GPT-4 and GPT-3.5. The scoring rubric of the task requires satisfying only one component, which is '"When the water is heated, water particles move faster (or increase in kinetic energy")." Both models generated responses indicating that "transferring heat energy changes the movement of the water" could be regarded as the component specified in the rubric. It is also notable that models corrected a misspelled word in the student's answer ('movment' → 'movement') even without related instruction. However, while GPT-4 strictly followed the structure of the rubric and example human scoring, GPT-3.5 did not. Although GPT-3.5 identified COMPONENT A, it ranked this student response as 'Beginning.' Further research has to be conducted to uncover why GPT-3.5 behaves this which results in less accurate outcomes. Figure 5 shows the difference of GPT-4 and GPT-3.5 on a student answer on Task J6_2.

# 5 Discussion

## 5.1 *LLM Yields Explainable Automatic Scoring*

Although research has shown the significant potential of AI in automatic scoring (Zhai, Haudek, et al., 2020), educators are concerned about the transparency and ethics of AI outcomes. Researchers have constantly called for establishing interpretability and explainability of AI applications and uncovering the "black box," so that AI applications can be embraced by general users (Du, Liu, & Hu, 2019). Our research suggests that LLMs can potentially tackle this issue. As shown in Figure 4 and Figure 5, GPT family produced natural language responses to the automatic scoring prompts of student-written answers to science items (RQ 1). Specifically, GPT-4 and GPT-3.5's responses identified the part of sentences in student-written answers, which uncovered the "black box" in scoring. Further, we found that users can instruct GPT's response style by instructing it to use prompt components.

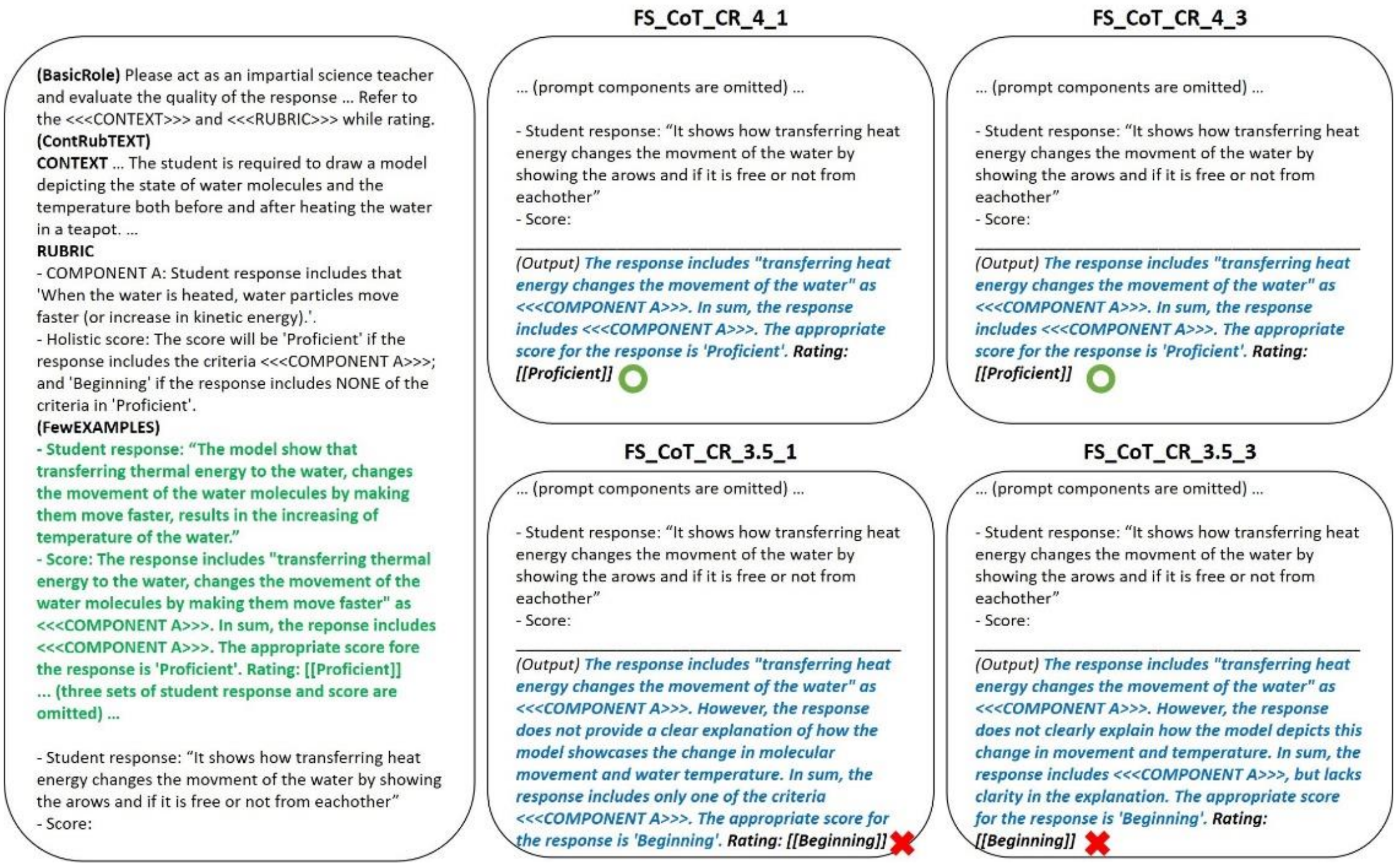

**Fig. 5** Comparison of the Answers from GPT-4 and GPT-3.5 (Task J6_2)

Technically, our results suggest the feasibility of LLMs for generating interpretable and explainable scores by leveraging prompt engineering. Interpretability has been a challenge in deploying AI applications in sectors of society, so researchers developed many strategies to unpack the "black box," such as placing additional layer with interpretable constraints in the model or extracting post-hoc features from the model (Du et al., 2019). These strategies are found effective but need substantial professional knowledge to develop and deploy, which is usually beyond the reach of general educators.

Practically, we found that GPT-4 and GPT-3.5 can generate natural language explanations using chatbot-like functions, which are effective for broader users, including teachers. Particularly, when two researchers with expertise in science education reviewed the responses from GPT based on the scoring rubrics (Figures 4-5), they found that response components extracted by GPT match the scoring rubric components, indicating plausibility. Further, we found that CoT yielded explicit and transparent scoring outcomes. This opens the essential possibility for human users to check and correct how scoring machines score student responses, which was impossible in previous automatic scoring systems. Differently speaking, if there are any issues related to ethics and bias in the LLM-based scoring models, it could be found in the CoT steps and fixed instructing the LLM with appropriate prompts, which sheds light to 'black-box' systems used for education. Further, teachers could also leverage LLMs to develop scoring rubrics, draft their evaluation of student responses, before they review, revise, and finalize their assessment, which could dramatically expediate the scoring process.

These advantages ease the efforts of developing fancy algorithms to uncover the interpretability of automatic scoring, evidencing a milestone improvement. Therefore, the adoption of LLMs could make a substantial change in automatic scoring research, making previous 'black-box' models explainable and user-friendly. This could facilitate the integration of AI technologies for automatic scoring even for classroom settings in the future so that teachers and researchers have a powerful tool to realize real-time feedback based on formative assessment.

### 5.2 *Zero-shot Versus Few-shot Learning*

This study demonstrated the high accessibility of prompt-based automatic scoring to broader populations other than professional developers. We found that the prompt engineering (e.g., zero-shot, few-shot learning) techniques are less data-demanding and labor-intensive, and thus are promising to transform the automatic scoring paradigm. Prior research on AI-based assessments usually needs a large training corpus to develop scoring algorithmic models, which is cost- and time-consuming (Zhai, Yin, Pellegrino, Haudek, & Shi, 2020). By using prompt engineering, we only need a small number of human-scored data to validate and test the prompts. Moreover, traditional AIbased assessments usually demand coding skills to create algorithmic models, which was beyond the reach of general educators. In this study, we leveraged GPT and prompt engineering to score student written responses without using any programming or strenuous model training. This advance sharply distinguishes this approach from the previous text classification models that mainly used complex machine learning algorithms without explanation.

Our findings suggest that prompting strategies are beneficial to improve zero-shot and few-shot scoring. In their prior research, Wu et al. (2023) reported scoring accuracy that needs improvement. By experimenting with various prompting strategies, we found that GPT-4 and GPT-3.5 could automatically score student-written responses with improved scoring accuracy. We started from the baseline prompts, such as ZS_noCoT and FS_noCoT, and furthered the investigation with CoT prompt components for the zero/few-shot learning approach. Our research suggests that few-shot learning shows 7.48% higher accuracy than zero-shot learning on average 4. These improvements suggest that few-shot learning is promising for automatic scoring tasks with significantly less human effort.

Compared with zero-shot learning, we found that few-shot learning leveraged examples provided and CoT. As shown in Figures 3-4, few-shot prompts tended to produce shorter and more structured responses compared to zero-shot learning, as like human evaluator's scoring examples. Particularly, when CoT was introduced to the prompts, GPT followed CoT reasoning steps to score students' answers. This structure-coercing effect of few-shot learning can be the main reason it improves scoring accuracy, imitating human scorers' way of thinking.

### 5.3 *Chain-of-Thought Paired with Contextual Instructions Contributes to Automatic Scoring*

Aligning with prior research (Kojima et al., 2022; Wei et al., 2022), our findings indicate that CoT is an efficient strategy for improving prompt engineering, particularly for the automatic scoring performance; yet CoT cannot work in isolation. CoT can be effective only when paired with the problem context and scoring rubrics. That is, we found that simply including CoT without the contextual instructions contributed limited to scoring model performance (Table 4), which somehow diverged from the examples (e.g., arithmetic, commonsense reasoning, letter concatenation) presented in the original CoT papers. This difference may be due to the complexity of automatic scoring of student-written responses, as compared to the example tasks presented in the papers that suggested CoT methods. This finding implies that the general reasoning elicited by CoT in LLMs has limited to improve automatic scoring (Wei et al., 2022), but the domain-specific or task-specific reasoning is critical to increasing scoring accuracy.

Consistent with the findings above, this study contributes to the literature on CoT prompting by introducing the WRVRT approach. WRVRT employs an iterative procedure to develop prompts that include multiple components. Specifically, we have to first specify the role of GPT by using the prompt "act as an impartial science teacher" (*BasicRole*). It leverages the few-shot learning examples, item contexts, and scoring rubrics and incorporates CoT to efficiently guide GPT to assign scores to students' written responses. We would analogize this role of guideline as 'a chain mold', which makes the chain of reasoning aligned and prepared to be used.

In addition, our qualitative analysis also revealed the processes of GPT in improving scoring accuracy when leveraging the WRVRT approach. We found that the model performance increases because the category-wise accuracy increases (Table 5). That is, scoring accuracy in some scoring categories increased, while decreased on others, and the overall performance is increased. This pattern was identical both in zero-shot and few-shot learnings. We suspect that this ability may specifically address unbalanced data issues, but future research should further unpack the mechanisms.

### 5.4 *GPT-4 versus GPT-3.5 by Hyperparameter*

Our results suggest that GPT-4 with single-call greedy sampling seems to be the best strategy for automatic scoring compared to GPT-3.5 with other strategies. Table 6 illustrates the superior performance of GPT-4 compared to GPT-3.5, evidenced in both the ensemble voting and single-call strategies. This result highlights that GPT-4's enhanced reasoning abilities are beneficial for automated grading systems. Therefore, it is recommended to use GPT-4 rather than GPT-3.5 in the automatic scoring of student-written responses concerning scoring accuracy. However, educators may also have to consider the availability of resources, given that GPT-4 API call is 30 times more expensive than GPT-3.5 as of November 2023.

We also found that the voting strategy may be only beneficial when using lower computation LLMs. The voting strategy is more resource-demanding and thus deserves investigations on its effectiveness. In this study, we found that the voting strategy with GPT-4 or 3.5 did not largely improve the average performance for all the tasks compared to using the single-calls strategy. Particularly, GPT-3.5 benefited from the voting strategy in majority tasks, although the improvement was minor. However, the voting strategy with GPT-4 seems less productive compared to the single-call greedy-sampling approach. This result suggests that the voting strategy is primarily aimed at reducing uncertainty in grading predictions for less advanced LLMs. The high confidence of GPT-4 in predictions marks a distinct behavior from GPT-3.5. However, in general, users need to balance the usability and cost, because the ensemble voting strategy requires three times additional computations while offering limited improvement in the model accuracy.

## 6 Conclusions

This study examined the affordance of GPT-4, equipped with CoT, on the automatic scoring of students' written explanations to science questions. The research findings underscore the feasibility of using LLMs to not only execute scoring tasks with high efficiency but also provide explainable and interpretable outcomes, which is vital in the context of educational assessments. Our investigation into the comparative performance of zero-shot and few-shot learning prompts revealed a marked improvement in scoring accuracy with the application of few-shot learning (12.6%). This advancement indicates a few-shot learning as a promising direction for automatic scoring tasks, reducing the need for extensive human input while maintaining high accuracy levels. Additionally, the CoT prompting strategy, especially when paired with contextual item stems and rubrics, proved to be a significant contributor to scoring accuracy (13.44% increase for zero-shot; 3.7% increase for few-shot). The study was conducted under a novel approach WRVRT, which was found to facilitate a more balanced accuracy across different proficiency categories, highlighting the importance of domain-specific reasoning in enhancing the effectiveness of LLMs in scoring tasks.

The study also suggests that GPT-4 demonstrated superior performance over GPT3.5 in various scoring tasks, showing 8.64% difference. The study revealed that the single-call strategy with GPT-4, particularly using greedy sampling, outperformed other approaches, including ensemble voting strategies. This finding suggests that the advanced reasoning abilities of GPT-4 are more conducive for automated scoring systems, offering greater reliability and accuracy. The nuanced understanding of the interaction between model capacity and voting strategy, alongside the exploration of the cost versus usability trade-offs, adds a practical dimension to the research, making it highly relevant for educators and researchers seeking to integrate AI technologies into their instructional and assessment practices.

Despite of the potential documented in this study, future research should continue improving the model capacity to increase the automatic scoring performance. While the scoring accuracy estimated in this study spanned .5885-.9083, the parental study that used the equivalent dataset to train and test the ensemble automatic scoring machine shows an accuracy spanning .86-.94, which is higher than the prompt engineering approach (Zhai, He, & Krajcik, 2022). This may be because the ensemble approach fine-tuned the parameters for specific tasks, while the zero- or few-shot learning approaches using pre-trained LLM are more generic. Therefore, users have to balance efficiency and accuracy for specific assessment purposes. Studies should further explore novel and sophisticated prompt engineering for LLMs to advance automatic scoring, delve into the nuance of students' thinking. In this regard, while this study focused on the final label of student-written responses for model performance, analytic scoring approach for each component of scoring rubric, which will correspond to CoT process, could be further studied with appropriate dataset. Also, the characteristics of items and scoring rubrics need to be investigated further to improve CoT outcomes.

## Acknowledgement

This study secondary analyzed a dataset in the publication (Zhai, He, & Krajcik, 2022). The authors are grateful for the team (e.g., Zhai, He, and Krajcik) as well as those colleagues contributed to the scoring (e.g., Jie Yang, Sisi Han, and Tingting Li) that contributed to the original study. The study was funded by the National Science Foundation(NSF) (Award no. 2101104). Any opinions, findings, conclusions, or recommendations expressed in this material are those of the author(s) and do not necessarily reflect the views of the NSF.

**Declaration of Interest statement**

None.

## Appendix 1: Comprehensive Example of Prompt Engineering Components (Task H4_3)

**BasicRole** (*: Concatenated only for ZS_CoT_CR and FS_CoT_CR)
Please act as an impartial science teacher and evaluate the quality of the response provided by a middle school student to a science item displayed below. Begin your evaluation by providing a short explanation. Be as objective as possible. After providing your explanation, you must classify the response on a scale of 'Beginning,' 'Developing,' and 'Proficient' by strictly following this format: "[[rating]]," for example: "Rating: [[Beginning]]." *(Refer to the <<<CONTEXT>>>and <<<RUBRIC>>>while rating).

**ContRubTEXT**

CONTEXT
Simone took a hot shower and wondered what would happen to the water vapor when it came in contact with a cold mirror. The task is to construct a model that illustrates the changes in water molecules from Simone's shower once they hit the cold mirror. This model should display the thermal energy and kinetic energy of the water molecules. The goal is to explain how the state of water vapor changes after it interacts with the cold mirror.

RUBRIC
- COMPONENT A: Student response includes an 'explanation that the substance changes its state from gas to liquid.'
- COMPONENT B: Student response includes that 'the change in state occurs because of a decrease in the particles' motion/kinetic energy.'

- Holistic score: The score will be 'Proficient' if the response includes ALL of the criteria <<<COMPONENT A>>>AND <<<COMPONENT B>>>; 'Developing' if the response includes at least ONE BUT NOT ALL of the criteria in 'Proficient;' and 'Beginning' if the response includes NONE of the criteria in 'Proficient.'

**FewEXAMPLES** (for FS_noCoT)

- Student response: "In water vapor, water molecules move fast and are far apart as a gas in the bathroom. When water molecules touch the cold mirror, thermal energy is transferred from the water molecules to the cold mirror. This causes the kinetic energy of the molecules of water vapor to decrease, the molecules to move slower as represented by the shorter arrows in the model, and the molecules to stay closer to each other like a liquid and form water droplets. So, the prediction is that the water vapor from Simone's shower (gas) will become water droplets (liquid)."
- Score: 'Proficient.' Rating: [[Proficient]]
- Student response: "the molecules are starting to get warmer, moving faster as they are turning into a gas."
- Score: 'Developing.' Rating: [[Developing]]
- Student response: "in the cold mirror, the water vapor is moving slower
- Score: 'Developing.' Rating: [[Developing]]
- Student response: "This shows that when the water vapor hits the mirror it can start to do evaporation this is what the picture represents."
- Score: 'Beginning.' Rating: [[Beginning]]

**FewEXAMPLES** (for FS_CoT and FS_CoT_CR)

- Student response: "In water vapor, water molecules move fast and are far apart as a gas in the bathroom. When water molecules touch the cold mirror, thermal energy is transferred from the water molecules to the cold mirror. This causes the kinetic energy of the molecules of water vapor to decrease, the molecules to move slower as represented by the shorter arrows in the model, and the molecules to stay closer to each other like a liquid and form water droplets. So, the prediction is that the water vapor from Simone's shower (gas) will become water droplets (liquid)."
- Score: The response includes "the water vapor ... (gas) will become water droplets (liquid)" as <<<COMPONENT A>>>. The response includes "the kinetic energy of ... water vapor to decrease" as <<<COMPONENT B>>>. In sum, the response includes ALL of the criteria <<<COMPONENT A>>>AND <<<COMPONENT B>>>. The appropriate score for the response is 'Proficient.' Rating: [[Proficient]]
- Student response: "the molecules are starting to get warmer moving faster as they are turning into a gas"
- Score: The response includes "turning into a gas" as <<<COMPONENT A>>>. The response does not include ¡¡¡COMPONENT B>>>. In sum, the response includes at least ONE BUT NOT ALL of the criteria <<<COMPONENT A>>> AND <<<COMPONENT B>>>. The appropriate score for the response is 'Developing.' Rating: [[Developing]]
- Student response: "In the cold mirror the water vapor is moving slower"
- Score: The response does not include <<<COMPONENT A>>>. The response includes "moving slower" as <<<COMPONENT B>>>. In sum, the response includes at least ONE BUT NOT ALL of the criteria <<<COMPONENT A>>>AND <<<COMPONENT B>>>. The appropriate score for the response is 'Developing.' Rating: [[Developing]]
- Student response: "This shows that when the water vapor hits the mirror it can start to do evaporation this is what the picture represents."
- Score: The response does not include <<<COMPONENT A>>>. The response does not include <<<COMPONENT B>>>. In sum, the response includes NONE of the criteria <<<COMPONENT A>>>AND <<<COMPONENT B>>>. The appropriate score for the response is 'Beginning.' Rating: [[Beginning]]

## Appendix 2: Overall Model Performance Metrics

- ZS: Zero-Shot, FS: Few-Shot, CoT: Chain-of-Thought, CR: Problem Context and Rubric, KappaQW: Quadratic Weighted Kappa, Acc: Accuracy, Prof: Proficient, Dev: Developing, Beg: Beginning
- Method: Calling GPT-4 API once with temperature = 0 and top_p = 0.01

| Task | Method | Accuracy | Precision | Recall | F1 | KappaQW | Acc_Prof | Acc_Dev | Acc_Beg |
|---|---|---|---|---|---|---|---|---|---|
| R1_2 | ZS_noCoT | 0.6625 | 0.6741 | 0.6625 | 0.6568 | NA | 0.7917 | NA | 0.5333 |
| | ZS_CoT | 0.6458 | 0.6697 | 0.6458 | 0.6329 | NA | 0.8333 | NA | 0.4583 |
| | ZS_CoT_CR | 0.7583 | 0.7839 | 0.7583 | 0.7528 | NA | 0.9083 | NA | 0.6083 |
| | FS_noCoT | 0.7833 | 0.7973 | 0.7833 | 0.7808 | NA | 0.675 | NA | 0.8917 |
| | FS_CoT | 0.75 | 0.8196 | 0.75 | 0.7356 | NA | 0.5167 | NA | 0.9833 |
| | FS_CoT_CR | 0.7625 | 0.8204 | 0.7625 | 0.7513 | NA | 0.55 | NA | 0.975 |
| J2_2 | ZS_noCoT | 0.6417 | 0.6744 | 0.6417 | 0.624 | NA | 0.8583 | NA | 0.425 |
| | ZS_CoT | 0.6417 | 0.6848 | 0.6417 | 0.6194 | NA | 0.8833 | NA | 0.4 |
| | ZS_CoT_CR | 0.8458 | 0.847 | 0.8458 | 0.8457 | NA | 0.875 | NA | 0.8167 |
| | FS_noCoT | 0.7958 | 0.8034 | 0.7958 | 0.7945 | NA | 0.7167 | NA | 0.875 |
| | FS_CoT | 0.8375 | 0.8616 | 0.8375 | 0.8347 | NA | 0.7083 | NA | 0.9667 |
| | FS_CoT_CR | 0.8792 | 0.8805 | 0.8792 | 0.8791 | NA | 0.85 | NA | 0.9083 |
| H4_2 | ZS_noCoT | 0.3613 | 0.5087 | 0.3975 | 0.3545 | 0.2791 | 0.2583 | 0.725 | 0.2091 |
| | ZS_CoT | 0.371 | 0.5061 | 0.4095 | 0.3601 | 0.2525 | 0.275 | 0.7625 | 0.1909 |
| | ZS_CoT_CR | 0.5935 | 0.6542 | 0.5994 | 0.5898 | 0.5806 | 0.6833 | 0.6875 | 0.4273 |
| | FS_noCoT | 0.5581 | 0.5414 | 0.539 | 0.5383 | 0.4701 | 0.6 | 0.3625 | 0.6545 |
| | FS_CoT | 0.5774 | 0.5973 | 0.5735 | 0.5681 | 0.5583 | 0.475 | 0.5 | 0.7455 |
| | FS_CoT_CR | 0.5452 | 0.6558 | 0.5713 | 0.5504 | 0.5004 | 0.4833 | 0.8125 | 0.4182 |
| H4_3 | ZS_noCoT | 0.4722 | 0.556 | 0.4722 | 0.4429 | 0.4258 | 0.5417 | 0.7417 | 0.1333 |
| | ZS_CoT | 0.5111 | 0.6096 | 0.5111 | 0.4832 | 0.4835 | 0.6083 | 0.7667 | 0.1583 |
| | ZS_CoT_CR | 0.6333 | 0.698 | 0.6333 | 0.6298 | 0.6276 | 0.775 | 0.7083 | 0.4167 |
| | FS_noCoT | 0.5917 | 0.566 | 0.5917 | 0.5589 | 0.6111 | 0.7083 | 0.225 | 0.8417 |
| | FS_CoT | 0.6806 | 0.7013 | 0.6806 | 0.6813 | 0.6667 | 0.5667 | 0.6833 | 0.7917 |
| | FS_CoT_CR | 0.6667 | 0.6971 | 0.6667 | 0.6734 | 0.6831 | 0.6833 | 0.675 | 0.6417 |
| J6_2 | ZS_noCoT | 0.6583 | 0.7332 | 0.6583 | 0.6285 | NA | 0.9417 | NA | 0.375 |
| | ZS_CoT | 0.6458 | 0.7479 | 0.6458 | 0.6052 | NA | 0.9667 | NA | 0.325 |
| | ZS_CoT_CR | 0.6792 | 0.8046 | 0.6792 | 0.6424 | NA | 1 | NA | 0.3583 |
| | FS_noCoT | 0.7833 | 0.7885 | 0.7833 | 0.7824 | NA | 0.7167 | NA | 0.85 |
| | FS_CoT | 0.825 | 0.8343 | 0.825 | 0.8238 | NA | 0.7417 | NA | 0.9083 |
| | FS_CoT_CR | 0.9083 | 0.9094 | 0.9083 | 0.9083 | NA | 0.8833 | NA | 0.9333 |
| J6_3 | ZS_noCoT | 0.4962 | 0.4677 | 0.3861 | 0.3631 | 0.186 | 0.2583 | 0.8 | 0.1 |
| | ZS_CoT | 0.5038 | 0.4728 | 0.3917 | 0.3715 | 0.1824 | 0.275 | 0.8 | 0.1 |
| | ZS_CoT_CR | 0.5885 | 0.5096 | 0.4528 | 0.4135 | 0.3467 | 1 | 0.2583 | 0.1 |
| | FS_noCoT | 0.45 | 0.4551 | 0.4917 | 0.4132 | 0.3309 | 0.5333 | 0.3417 | 0.6 |
| | FS_CoT | 0.2385 | 0.3806 | 0.4222 | 0.2375 | 0.2351 | 0.3417 | 0.025 | 0.9 |
| | FS_CoT_CR | 0.4231 | 0.5061 | 0.5 | 0.35 | 0.4045 | 0.75 | 0.05 | 0.7 |